\documentclass[conference]{IEEEtran}
\IEEEoverridecommandlockouts
\usepackage{cite}
\usepackage{amsmath,amssymb,amsfonts}
\usepackage{algorithmic}
\usepackage{graphicx}
\usepackage{textcomp}
\usepackage{xcolor}
\def\BibTeX{{\rm B\kern-.05em{\sc i\kern-.025em b}\kern-.08em
    T\kern-.1667em\lower.7ex\hbox{E}\kern-.125emX}}
\begin{document}

\title{Efficient Visualization of Neural Networks with Generative Models and Adversarial Perturbations \\

}

\author{\IEEEauthorblockN{Athanasios Karagounis}
\IEEEauthorblockA{\textit{School of Science} \\
\textit{Department of Digital Industry Technologies}\\
GR34400, Psachna, Greece \\
akaragun@gs.uoa.gr}

}

\maketitle

\begin{abstract}
This paper presents a novel approach for deep visualization via a generative network, offering an improvement over existing methods. Our model simplifies the architecture by reducing the number of networks used, requiring only a generator and a discriminator, as opposed to the multiple networks traditionally involved. Additionally, our model requires less prior training knowledge and uses a non-adversarial training process, where the discriminator acts as a guide rather than a competitor to the generator. The core contribution of this work is its ability to generate detailed visualization images that align with specific class labels. Our model incorporates a unique skip-connection-inspired block design, which enhances label-directed image generation by propagating class information across multiple layers. Furthermore, we explore how these generated visualizations can be utilized as adversarial examples, effectively fooling classification networks with minimal perceptible modifications to the original images. Experimental results demonstrate that our method outperforms traditional adversarial example generation techniques in both targeted and non-targeted attacks, achieving up to a 94.5\% fooling rate with minimal perturbation.
This work bridges the gap between visualization methods and adversarial examples, proposing that fooling rate could serve as a quantitative measure for evaluating visualization quality. The insights from this study provide a new perspective on the interpretability of neural networks and their vulnerabilities to adversarial attacks.
\end{abstract}


\section{Introduction}
The rapid advancements in deep learning have driven significant progress in various domains, including computer vision, natural language processing, and autonomous systems. However, as deep neural networks (DNNs) become more complex, the need for interpretability and visualization of these models has become critical. Understanding what a neural network learns and how it makes decisions is essential, especially in safety-critical applications such as medical diagnostics, self-driving cars, and security systems.

Recent methods for visualizing DNNs, such as those proposed by Nguyen et al. [10], rely on generating images that represent the preferred inputs of neurons through adversarial training. While these techniques have demonstrated promising results, they often suffer from overly complex architectures and reliance on multiple pre-trained networks. Additionally, many of these methods involve adversarial training processes that are computationally expensive and difficult to interpret.

In this paper, we propose a simplified yet effective method for deep visualization using generative networks. Our model reduces the architectural complexity by employing only two networks: a generator and a discriminator. Unlike existing methods, which use multiple networks for encoding, generating, and comparing images, our approach streamlines the process while maintaining high-quality visual outputs. Furthermore, our model only pre-trains the discriminator on real datasets, while the generator is trained under the supervision of the discriminator, without the need for adversarial training.

An additional focus of this work is the relationship between visualization techniques and adversarial examples. Adversarial examples are subtly modified inputs designed to mislead neural networks into incorrect predictions. These examples reveal the vulnerabilities of DNNs and raise important questions about their robustness. We show that our visualization results can be utilized as effective adversarial perturbations, capable of fooling DNNs with high accuracy. Our experiments demonstrate that this method can achieve a fooling rate of up to 94.5\% with minimal perturbation, illustrating both the power and the interpretability of our approach.

This paper is organized as follows: Section 2 introduces related work, including generative adversarial networks and adversarial examples. Section 3 outlines the architecture of our proposed model, and Section 4 details the training process. Section 5 presents experimental results and analysis. Finally, Section 6 concludes with a discussion of the implications and future directions of our research.

\begin{figure*}
 \centering
 \includegraphics[width=0.70\textwidth]{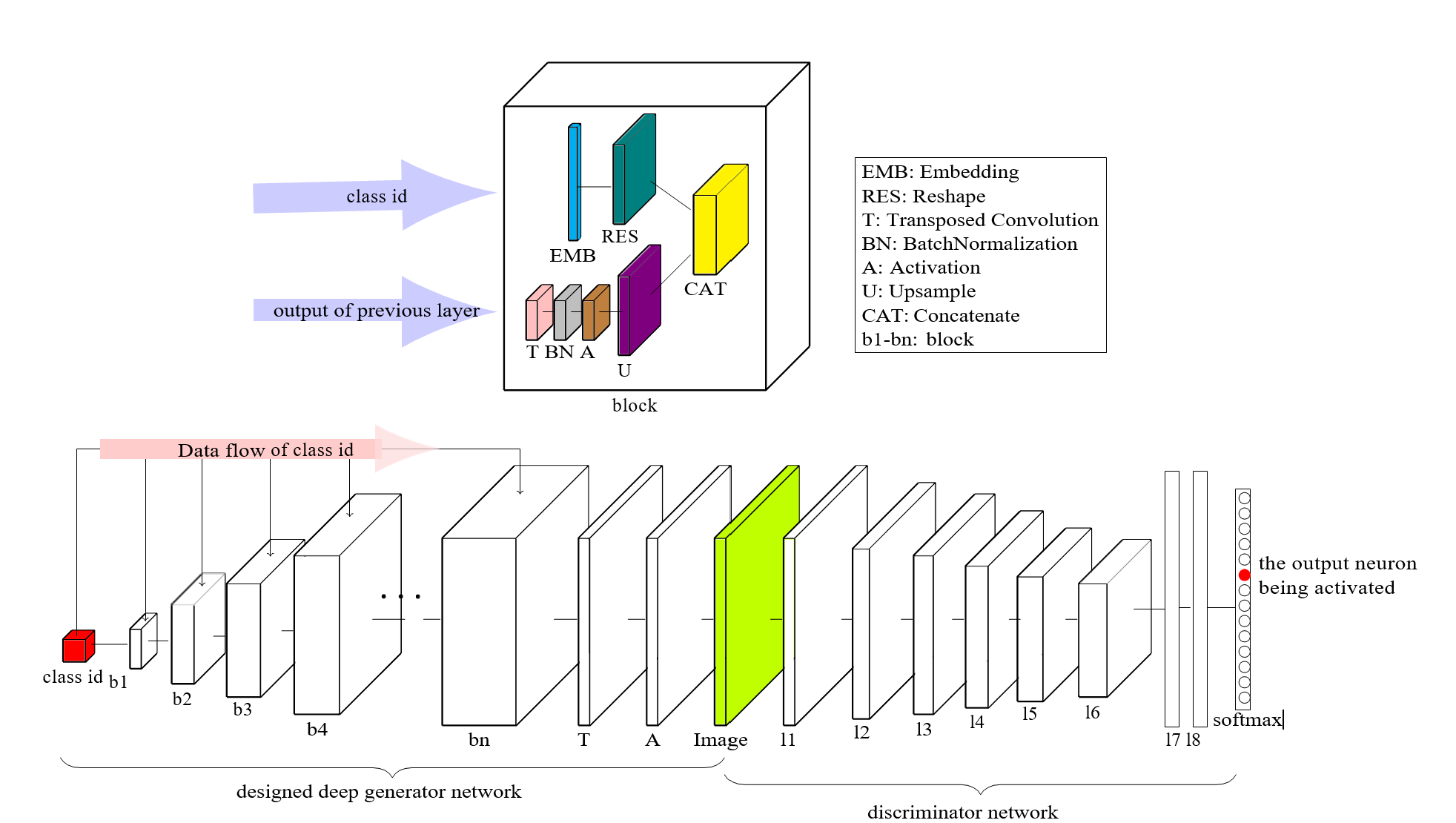}
 \caption{The illustration of the proposed model. Bottom row represents the overall structure. The proposed generative network is shown on the left and the classification network that needs to be visualized is shown on the right. Given a class id (red cube) as input, the generator can generate an image. The discriminator makes a prediction on the generated image. The loss is measured by the distance between input class id (red cube on the left) and predicted class id (red circle on the right). We freeze the discriminator’s weights and only train the generator. The top row represent our designed block to encode the input class id information (pink arrow in the bottom row) to layers of different depth for generator network.}
 \label{fig1}
\end{figure*}

\section{Related Work}
\subsection{Generative Adversarial Networks}
As claimed in \cite{nguyen2016synthesizing}, their model used four convolutional networks: an encoder network E, a generator network G, a ``comparator'' network C and a discriminator network D. Our model only uses two networks: a generator and a discriminator. Noth the E and
C have been trained on real image dataset. Both the G and D have been trained adversarially before. For our model, only the discriminator network has been pre-trained on the real 	dataset. Also, the training of our model is not adversarially, the discriminator only acts as a leader of generator.
The discriminator is 	used to judge whether the generated image is fake or not. For our method, the discriminator is used to classify images to $1000$ classes. Besides, they also used a ``comparator'' network 	to measure the difference between generated image and real image.  Our model has no
particular mechanism to constrain the interpretation of the generated images. Our goal is to obtain a model, and then use the model to generate
visualization images.  Their goal is to obtain different codes, and then use the codes to	generate visualization images.

\subsection{Adversarial Examples}

Adversarial examples \cite{goodfellow2014explaining} are a type of input data that have been modified very slightly with the intent of causing a machine learning classifier to misclassify the input. For example, given an image of a panda, an attacker can add a small perturbation that has been calculated to make the image be recognized as a gibbon with high confidence. This process is referred to as fooling the classifier. Most methods \cite{goodfellow2014explaining} for generating adversarial image examples are based on a specific image: that is, given an image with a specific, mistaken, targeted label, we can calculate a perturbation corresponding to that image. However, that perturbation may not be effective when added to another image. There is also a type of perturbation noise called universal perturbation \cite{moosavi2017universal}, which can be added to different images and cause the network to make mistakes \cite{amanatiadis2018understanding},\cite{sophokleous2022educational}; this method has no pre-defined incorrect label, and is therefore non-targeted.

Both of the above methods are optimization-based. In this paper, we find that our generated visualization results can easily be used as perturbations added to natural images to generate adversarial examples. More details will be provided in the experiment section.

\begin{figure*}
 \centering
 \includegraphics[width=0.8\textwidth]{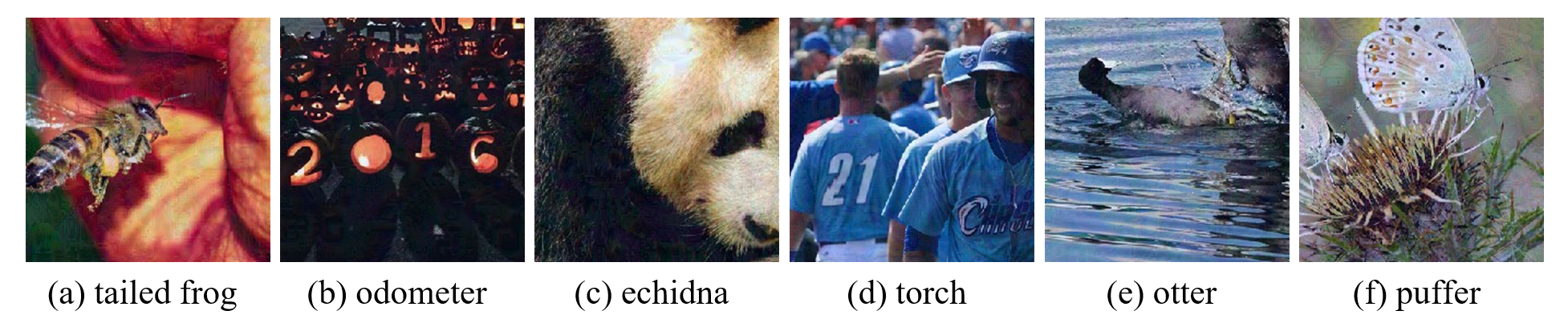}
 \caption{Adversarial examples and their corresponding mistaken labels.}
 \label{fig2}
\end{figure*}

\begin{figure*}
 \centering
 \includegraphics[width=0.8\textwidth]{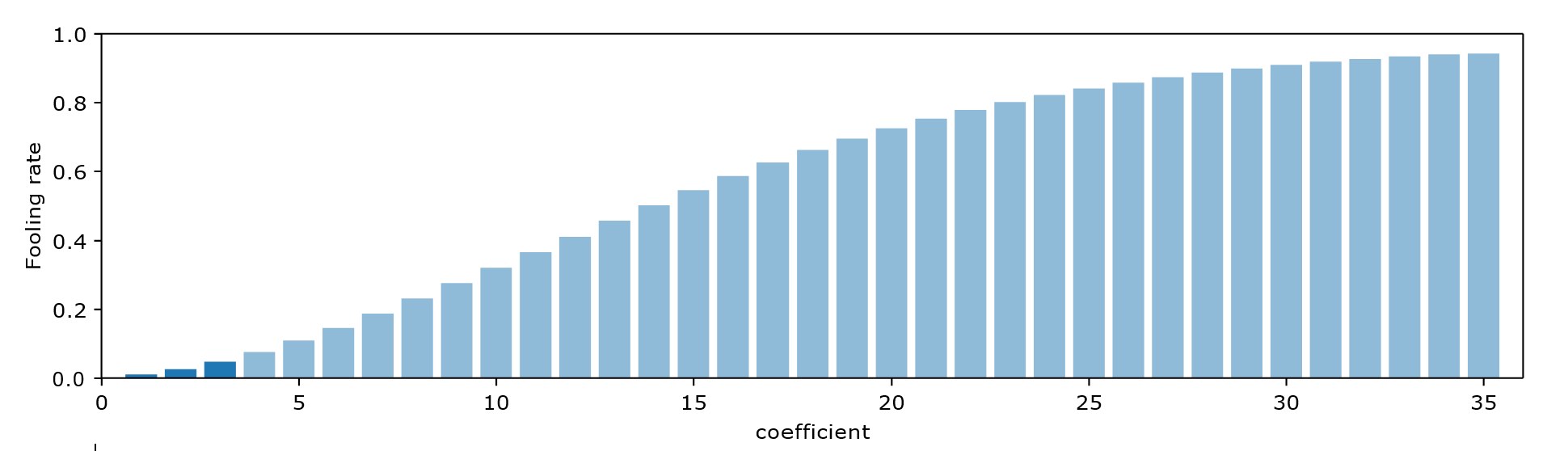}
 \caption{The relationship of multiplied coefficient and fooling rate.}
 \label{fig3}
\end{figure*}

\section{Proposed Method}
\subsection{Network Architecture}

The proposed visualization architecture is illustrated in Figure \ref{fig1}. It consists of two subnetworks: a generative network (left) and a discriminative network (right). The discriminator network is actually a typical classification network that needs to be visualized. In this paper, we use the CaffeNet \cite{jia2014caffe} and VGG16 net \cite{simonyan2014very} that has already been trained on the ImageNet dataset \cite{deng2009imagenet} to perform the experiment. Other image classification networks can also be easily visualized following the same procedure.
Layers for discriminator network:
\begin{itemize}
	\item l1-l6: Common image features extracting layers of classification network, it can be convolution, pooling or activation layers \cite{mountelos2019vehicle}. Outputs of these layers are 3D matrixes, containing 
	\begin{flushright}
	
	\end{flushright}
	feature maps with multiple channels.
	\item l7-l8: Fully connected layers. l7 is used convert the 3D matrix into 1D vectors. l8 is used to change the length of 1D vector to exactly the number of classes.
	\item softmax layer: An activation layer which can exponentially normalize the 1D vector to sum value of 1. It is common used as the output layer for the discriminator network. The output	value of each neuron represent the possibility for each class. If we want to visualize that	class, we will activate that corresponding neuron(maximize the corresponding possibility).
\end{itemize}

The generator network is used to output a visualization image corresponding to a class id as input. Common  image  generation  network  is  deconvolution  network \cite{zeiler2010deconvolutional},\cite{faniadis2020deep}. However, just  inputting  the label information into the first layer of the generator is hard for the model to converge, some input information of the first layer might get lost when it goes through the whole network because of vanishing gradient problem \cite{pascanu2013difficulty}.  To overcome this problem, the input information should be strengthened.  We learned from the skip-connected thought from ResNet \cite{he2016deep} and feed the input information to modules in different depths of the generative network. The block module(the first row of Fig. \ref{fig1} is designed to encode the class id information into dense matrix and combined the matrix with data of layers in different depth. This module is critical for the generator to be label directed.

\subsection{Training Process}

We use N to represent the number of classes of the classification network, x represents the input class label among N classes. For example, there are $1000$ classes in the ImageNet dataset [1], thus N  equals to $1000$ here and the length of the discriminator’s output vector is also $1000$.
Training steps:
\begin{itemize}
	\item Random choose class id $x$ between $0$ and $N$.
	\item Feed $x$ to the generator, then the generator generate an image $G(x)$.
	\item Feed the generated image to the image classification network D, then the network can produce a prediction vector $D(G(x))$.
	\item Calculate the categorical cross-entropy loss between the one hot encoding of initial class id (represented by $Gencode(x)$ and prediction possibility $vectorDout(G(x))$.
	\item Taking the two network as a whole, and using the above Loss, the gradient for each layer can be calculated, we freeze the discriminator and only update the weights of generator
\end{itemize}

Through the training procedure of the network, the above steps are repeated once and once, until the Loss reaches a small value. Then the generator is what we want, given an class id x, the generated image G(x) is the corresponding visualization image.Besides classification layer, our model can also be used to visualize the middle layers of networks. Each layer usually has several channels. In this paper, we take all the neurons of each channel as a whole and try to generate images that can activate all of the channel neurons. Our above model needs
 to be modified a little to realize this new task. For the discriminative network, we also use the pre-trained classification network. The output layer
      is not the last layer of the network but the layer we want to visualize. For each channel in a layer, the mean value of all the neuron activations on this channel is taken as the activation of the whole channel.  Therefore, for each layer, the number of activations is equal to the number of channels in this layer.  Then, we add a Softmax layer after the channel activation layer.  Different channels  are regarded as the previous different classes.  Compared with the model to visualize class labels, this model replaces the label indexes with channel indexes. Finally, the same generative network and training strategy are used to visualize different channels of the middle layers.

\section{Experimental Results}

In the experiment, we find that the generated image matrix can be used as perturbation on natural images.  For natural images, we use the dataset of NIPS2017 Adversarial Attacks and Defences Competition, which contains $1000$ images belonging to different labels. For the fooled classification network, we used the VGG16 network to generated perturbation and classify images before and after adding the perturbations. The values of generated matrix are between $[-1,1]$, we multiply the matrix with integer coefficient and add that to natural images.\par
Fooling images are shown in Fig. \ref{fig2}.  Each images represent natural images with perturbation generated by our method on it, with mistaken label showing below it.  The added perturbation  are hardly recognizable by human beings’ eyes, if you enlarge the PDF file, perhaps some detail can be       sensed. In the first image,the bee is wrongly classified as tailed frog; In the third image, the giant panda is wrongly classified as echidna. Results show that the perturbed image can make the network make totally wrong decision, and the wrong label is exactly the label of added visualized image. We call the perturbed images have fooled the neural network.\par
Since nowadays there are still no quantitative measure of visualization result \cite{amanatiadis2009survey}, most comparison are based on natural degree of images decided by human eyes. Based on the relationship between visualization and adversarial examples, we hope the fooling rate can be a good measure of visualization effectiveness. In Figure 3, we measured the relationship of multiplied coefficient and fooling rate.
The experiment is also taken on the NIPS 2017 Adversarial Attacks and Defenses Competition dataset. For each original image, the classifier predict an correct label $L_{correct}=argmax(D(I_{ori}))$, then   add a corresponding perturbation, for example, visualization matrix for label i, represented by Vi, then the prediction on perturbed image is $L_{perturbed}=argmax(D(I_{ori} + e   Vi))$,  if the $L_{correct}$  does not equal to the $L_{perturbed}$, we called it a successful fooling. For all the class $i$ in $N$ (the number of classes) and all the images in the dataset, we can calculate a mean value of fooling rate for this coefficient $e$. We test on $e$ between $1$ and $35$, and plot the corresponding fooling rate in Fig. \ref{fig3}, with the coefficient increasing, the fooling rate is also increasing. For the coefficient $10$, which means the maximum changed pixel value is $10$, that is no obvious to humans, the fooling rate is $0.3$. And for the coefficient $35$, with less than $0.138$ of pixel range $255$, the method can fool almost $94.5\%$ images.\par
The above experiments showed the effectiveness for our method to generate adversarial examples. Different from previous adversarial examples generation methods, the whole process does not rely on complex math formula to realize optimization \cite{nguyen2016synthesizing},\cite{ioffe2015batch}, and the results generated from our method can meet both the needs of being targeted and can taking effect on different images. This work make a bridge between visualization methods and adversarial examples, as they are both methods to interpret inner mechanism of deep neural networks. Also, we hope that the fooling rate can be a good measure of visualization quality.

\section{Conclusions}
In this paper, we introduced a novel approach to deep visualization using a generative network, offering significant improvements over existing methods. By reducing the architectural complexity and eliminating the need for adversarial training, our model provides an efficient yet powerful means of generating visualization images for neural networks. Our streamlined method, which uses only a generator and a discriminator, ensures that the visualization process remains both interpretable and computationally efficient.
We also explored the connection between visualization and adversarial examples, demonstrating that the visualizations produced by our model can be repurposed as adversarial perturbations. Our experiments showed that the generated adversarial examples can fool state-of-the-art image classification networks with an adequate fooling rate, even with minimal perturbation that is nearly imperceptible to the human eye. This finding underscores the dual purpose of our approach: not only does it enhance the interpretability of neural networks, but it also highlights their vulnerabilities to adversarial attacks.

\bibliographystyle{IEEEtran}
\bibliography{arxivbibtex}

\end{document}